\documentclass[]{fairmeta}
\usepackage[utf8]{inputenc} % allow utf-8 input
\usepackage[T1]{fontenc}    % use 8-bit T1 fonts
\usepackage{hyperref}       % hyperlinks
\usepackage{url}            % simple URL typesetting
\usepackage{booktabs}       % professional-quality tables
\usepackage{amsfonts}       % blackboard math symbols
\usepackage{nicefrac}       % compact symbols for 1/2, etc.
\usepackage{microtype}      % microtypography
\usepackage{xcolor}         % colors
\usepackage{xspace}         % colors
\usepackage{graphicx}
\usepackage{sidecap}
\usepackage{subcaption}
\usepackage{float}

\usepackage{xspace}
\makeatletter
\DeclareRobustCommand\onedot{\futurelet\@let@token\@onedot}
\def\@onedot{\ifx\@let@token.\else.\null\fi\xspace}
\def\eg{\emph{e.g}\onedot} 
\def\ie{\emph{i.e}\onedot} 
 
\def\etc{\emph{etc}\onedot}

\makeatother

\newcommand{\memu}{Imagine yourself\xspace}

\title{\memu: Tuning-Free Personalized Image Generation}

\author{Zecheng He$^\ast$}
\author{Bo Sun$^\ast$}
\author{Felix Juefei-Xu$^\ast$}
\author{Haoyu Ma}
\author{Ankit Ramchandani}
\author{Vincent Cheung}
\author{Siddharth Shah}
\author{Anmol Kalia}
\author{Harihar Subramanyam}
\author{Alireza Zareian}
\author{Li Chen}
\author{Ankit Jain}
\author{Ning Zhang}
\author{Peizhao Zhang}
\author{Roshan Sumbaly}
\author{Peter Vajda}
\author{Animesh Sinha$^\ast$}
\affiliation{GenAI, Meta}
\abstract{Diffusion models have demonstrated remarkable efficacy across various image-to-image tasks. In this research, we introduce \memu, a state-of-the-art model designed for personalized image generation. Unlike conventional tuning-based personalization techniques, \memu operates as a tuning-free model, enabling all users to leverage a shared framework without individualized adjustments. Moreover, previous work met challenges balancing identity preservation, following complex prompts and preserving good visual quality, resulting in models having strong copy-paste effect of the reference images. Thus, they can hardly generate images following prompts that require significant changes to the reference image, \eg, changing facial expression, head and body poses, and the diversity of the generated images is low. To address these limitations, our proposed method introduces 1) a new synthetic paired data generation mechanism to encourage image diversity, 2) a fully parallel attention architecture with three text encoders and a fully trainable vision encoder to improve the text faithfulness, and 3) a novel coarse-to-fine multi-stage finetuning methodology that gradually pushes the boundary of visual quality. Our study demonstrates that \memu surpasses the state-of-the-art personalization model, exhibiting superior capabilities in identity preservation, visual quality, and text alignment. This model establishes a robust foundation for various personalization applications. Human evaluation results validate the model's SOTA superiority across all aspects (identity preservation, text faithfulness, and visual appeal) compared to the previous personalization models.}

\date{\today}
\correspondence{\email{zechengh@meta.com}}

\begin{document}

\maketitle

{\let\thefootnote\relax\footnote{* Core contributors}
% \begin{abstract}
% \end{abstract}

\section{Introduction}\label{sec:intro}

% \felix{Just a small change regarding \cite. When used in the beginning of a sentence, we use \citep{wei2023elite}. When used in middle of sentence, we use \citep{wei2023elite}, which will put the xxx in a parenthesis.}

Large scale diffusion models have drawn significant attention. These models, trained on vast amounts of image-text pairs, showcase remarkable semantic understanding capabilities and are able to generate diverse, photo-realistic images based on textual prompts. Due to their unparalleled creative abilities, large-scale diffusion models have found applications across a spectrum of image-to-image tasks beyond the original text-to-image generation, \eg, image editing, image completion, style transfer, and controllable generation.

% ------------------------------------------
\begin{figure}[ht!]
% \centering
\includegraphics[width=0.97\linewidth]{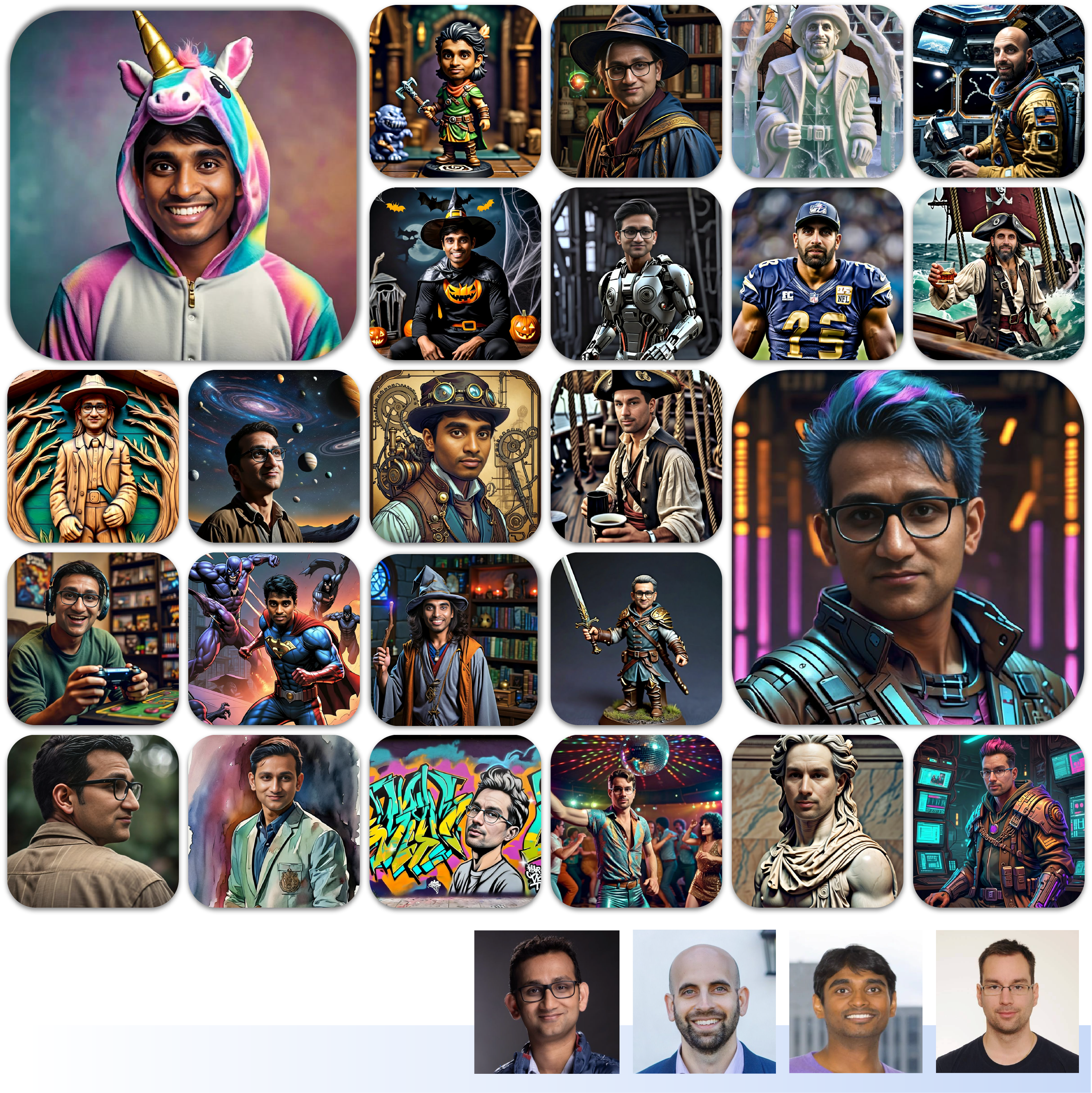}
\caption{Generated results for the four reference images (depicted below) using \memu. The single reference image is used to generate those subjects in novel poses and styles.} \label{img:intro:teaser}
\end{figure}
% The reference images are shown in Figure~\ref{img:intro:teaser_ref}
% ------------------------------------------

Personalized image generation techniques have gained significant attention alongside large-scale diffusion models. These methods focus on tailoring image generation to individual preferences or specific user characteristics. By incorporating customization into the generation process, these techniques aim to create images that are more relevant and appealing to the individual user. One line of research tunes a text-to-image model to incorporate the identity \citep{gal2022image,ruiz2023dreambooth,ruiz2023hyperdreambooth} with a few reference images. However, these methods are not efficient or generalizable as they require a different model to be tuned for each new user.

Recently, another effort has been proposed to obtain personalized diffusion models without subject-specific tuning. This direction of research extracts vision embedding from a reference images and inject it to the diffusion process \citep{wei2023elite, li2023photomaker, chen2023photoverse, ye2023ip, wang2024instantid, zhang2023adding, ostashev2024moa}. While the previous work in this direction can achieve a personalized model generalizable to all users, it usually comes with a strong over-fitting behavior, \ie, a copy-paste effect, to the reference image. Thus, they can hardly generate images following prompts that require significant changes to the reference image, \eg, change facial expression, head and body poses, hence the diversity of the generated images is low. As a result, these models cannot preserve identity while following complex prompts at the same time.

In this work, we propose \memu, a state-of-the-art model for personalized image generation without subject-specific fine-tuning. Unlike previous tuning-based personalization techniques which require tuning for each user, \memu is a tuning-free model where all subjects share a single model. We investigated the key components that lead to the quality improvement of \memu: \textbf{Identity preservation:} Trainable vision encoders with zero conv initialization, and masked vision embedding, \textbf{Visual quality:} multi-stage finetune and human-in-the-loop (HITL), \textbf{Text alignment:} Synthetic data and parallel attention. Meanwhile, we show that \memu outperforms the SOTA personalization models \citep{ye2023ip, wang2024instantid}, with significant margin in all aspects including identity preservation, visual quality, and text alignment, through large scale human evaluation. In particular, we won +27.8\% in text alignment compared to SOTA on complex prompts.

Our contributions can be summarized as follows: 
\begin{enumerate}
    \item We propose \memu, an innovative state-of-the-art model for personalized image generation. The proposed model can take any reference image as input for customized image generation and does not need tuning for each subject.
    \item \memu incorporates new components and shows significant improvements over the existing models: a new synthetic paired data generation mechanism to encourage image diversity, a fully parallel attention with three text encoders and a fully trainable vision encoder architecture to improve the text faithfulness, and a novel coarse-to-fine multi-stage finetuning methodology that gradually pushes the boundary of visual quality.
    \item We provide comprehensive qualitative and quantitative evaluation results compared to the state-of-the-art models. We provide human annotations on thousands of test examples as a golden standard to demonstrate the superior performance of \memu in all aspects, including identity preservation, prompt alignment, and visual appeal.
\end{enumerate}

% ------------------------------------------
% \begin{figure}[h]
% \centering
% \includegraphics[width=0.6\linewidth]{figures/teaser_v4b.pdf}
% \caption{Reference images for Figure~\ref{img:intro:teaser}.} \label{img:intro:teaser_ref}
% \end{figure}
% ------------------------------------------

% ------------------------------------------
\begin{table}[htbp]
  \centering
  \scriptsize
  \caption{Quantitative evaluation results of \memu vs. the SOTA control-based model and the SOTA adapter-based model under head-to-head human evaluation setting.}
  \begin{tabular}{c|ccc|ccc}
    \toprule
    ~ & \multicolumn{3}{c}{\textbf{Head-to-head} (win rate)} & \multicolumn{3}{c}{\textbf{Head-to-head} (win rate)} \\ \midrule
    \textbf{Metrics}        & \textbf{SOTA control-based model} & \textbf{\memu} & \textbf{Tie} & \textbf{SOTA adapter-based model} & \textbf{\memu} & \textbf{Tie} \\ \midrule
    Prompt Alignment        & 1.2\% & \textbf{46.3\%} & 52.6\% & 1.6\% & \textbf{32.4\%} & 66.0\% \\
    Identity Preservation   & \textbf{15.1\%*} & {3.2\%} & 81.7\% & 3.8\% & \textbf{5.5}\%  & 90.7\% \\
    Visual Appeal           & 11.5\% & \textbf{31.6\%}  & 57.0\% & 3.3\% & \textbf{4.2}\%  & 92.5\% \\
    \bottomrule
  \end{tabular} \label{tab:hev_h2h_x2}
\end{table}
% ------------------------------------------

{\let\thefootnote\relax\footnote{* We observed the SOTA control-based model is better in identity preservation than \memu, due to its hard copy-pasting of the reference image at the center of the image, resulting in unnatural images despite the high identity metric.}

\section{Related Work}\label{sec:related}

\subsection{Text-to-Image Diffusion Models}

Text-to-image diffusion models represent a cutting-edge paradigm in the domain of deep learning, captivating researchers with their capacity to translate textual descriptions into vibrant visual representations. At their core, these models operate through an iterative refinement process, wherein an initial noise vector is progressively denoised based on text prompts, ultimately yielding the desired image output. A common practice is first translating images to a latent space and denoising in that space. Stable Diffusion \citep{rombach2022high} and its variants SDXL \citep{podell2023sdxl}, Stable Diffusion Turbo \citep{sauer2023adversarial}, and Stable Diffusion-3 \citep{esser2024scaling} follow this path by increasing the model size, distilling a large model to fewer denoise steps, and leveraging new transformer architectures, respectively.

%This iterative journey involves multiple stages, each meticulously designed to enhance the image's quality while faithfully preserving its semantic context. With their ability to generate high-fidelity images from text, text-to-image diffusion models have emerged as a promising avenue in computer vision and artificial intelligence research, offering transformative potential across a spectrum of applications, from creative image synthesis to practical image editing tasks

\subsection{Tuning-based Personalization Models}

Diffusion-based personalized image generation has indeed garnered increased attention in recent times. This approach involves leveraging diffusion models to generate high-quality and personalized images based on a given input or set of inputs. Technically two streams of personalization models have been proposed. One stream of research tunes a text-to-image model to incorporate the identity. Textual Inversion \citep{gal2022image} finetunes a special text tokens for the new identity. DreamBooth \citep{ruiz2023dreambooth} leverages a few images from the same person as reference and a special text token to represent the identity. To accelerate the finetuning process, LoRA \citep{hu2021lora} only tunes a light-weight low-rank adapter rather than the whole diffusion model. HyperDreambooth \citep{ruiz2023hyperdreambooth} further predicts the initial weights of LoRA from the reference images. However, a major drawback of tuning-based personalization model is that the finetuned model becomes specific for the corresponding identity, and cannot be generalized to new identities. Furthermore, tuning for each user is costly and introduces a long waiting time.

\subsection{Tuning-free Personalization Models}

To overcome the limitations of the tuning-based method, another line of research focuses on one generalized model without identity-specific finetuning. This direction of work extracts vision embedding from the reference image and injects it to the diffusion process. ELITE \citep{wei2023elite} extracts vision features from reference image and converts it to the text-embedding space through a local and a global mapping. PhotoMaker \citep{li2023photomaker} merges the vision and text tokens and replaces the original text tokens for cross-attention. PhotoVerse \citep{chen2023photoverse} incorporates an image adapter and a text adapter to merge the vision and language tokens, respectively. IP-Adapter-FaceID-Plus \citep{ye2023ip} leverages face embedding and clip vision encoder for identity preservation. InstantID is a control-based method that \citep{wang2024instantid} adds ControlNet \citep{zhang2023adding} to further control the pose and facial expression. MoA \citep{ostashev2024moa} proposes a mixture of attention architecture to better fuse the vision reference and the text prompts.

\section{Method} \label{sec:method}

Our proposed \memu takes a single face image of a specific subject, and generates visually appealing personalized images guided by text prompts. Our method can follow complex prompt guidance and generate images with diverse head and body poses, expressions, style, and layout.
%layout (\animesh{What does layout mean?})\felix{layout might mean image composition, where each main object is in the picture}.

To push the boundaries of personalized image generation, our approach begins by identifying three key facets crucial to eliciting a satisfying human visual experience: identity preservation, prompt alignment, and visual appeal (Section \ref{sec:method:overview}). We then introduce novel techniques tailored to enhance each of these aspects. Specifically, we propose a novel synthetic paired data generation mechanism (Section \ref{sec:method:synpair}), new fully parallel architecture that incorporates three text encoders and a trainable vision encoder for optimizing identity preservation and text-alignment (Section \ref{sec:method:arch}), a novel coarse-to-fine multi-stage finetuning methodology designed to progressively enhance visual appeal, thereby pushing the visual appeal boundary of generated images. (Section \ref{sec:method:finetune}). Finally, we demonstrate \memu is generalizable to multi-subjects personalization in Section \ref{sec:method:multi_people}.

\subsection{Preliminary}

Text-to-Image diffusion models gradually turn a noise $\epsilon$ to a clear image $x_0$. While the diffusion process can happen in the pixel space \citep{ramesh2022hierarchical, saharia2022photorealistic}, a common practice is to have latent diffusion models (LDM) perform diffusion process in a latent space $\mathbf{z}=\mathcal{E}(x_0)$. During training, the LDM models optimize the reconstruction loss in the latent space:

\begin{equation}
    \mathcal{L}_\mathrm{diffusion} = \mathbb{E}_{z\sim\mathcal{E}(x_0),\epsilon\sim\mathbf{N}(0,1)} \|\epsilon - \epsilon_{\theta}(\mathbf{z}_t), t \|_2^2
\end{equation}

where $\mathcal{L}_\mathrm{diffusion}$ is the diffusion loss. $\epsilon_\theta$ represents the diffusion model. $\mathbf{z}_t$ is the noised input to the model at timestep $t$.

It is a common practice to use text or other condition signals $\mathbf{C}$ to guide the diffusion process. Thus, the conditioned diffusion process generates images following the condition signals. Usually, the text condition is incorporated with the diffusion model through cross-attention mechanism:

\begin{equation}
    \mathrm{Attn}(\mathbf{Q}, \mathbf{K}, \mathbf{V}) = \mathrm{softmax}\left(\frac{\mathbf{Q}\mathbf{K}^T}{\sqrt{d}}\right)\mathbf{V}
\end{equation}

where $\mathbf{K}=\mathbf{W}_K C$, $\mathbf{V}=\mathbf{W}_V C$ represents transforms that map the condition $C$ to the cross-attention key and values. $\mathbf{Q}=\mathbf{W}_Q \phi(x_t)$ represents the hidden state of the diffusion model.

\subsection{Overview} \label{sec:method:overview}

Figure \ref{img:method:overview} provides an illustration of the proposed model architecture. The key of using diffusion models for personalized image generation is incorporating the reference identity as an additional control signal to the diffusion model. We propose to extract the identity information from the reference image through a trainable clip patch encoder. The identity vision signal is then added to the text signals through a parallel cross attention module. To better preserve the high visual quality of the foundation model, we leveraged low-rank adapters (LoRA) to freeze the self-attention and text cross-attention modules while only fine-tuning the adapters.
% ------------------------------------------
\begin{figure}[h]
\includegraphics[width=\linewidth]{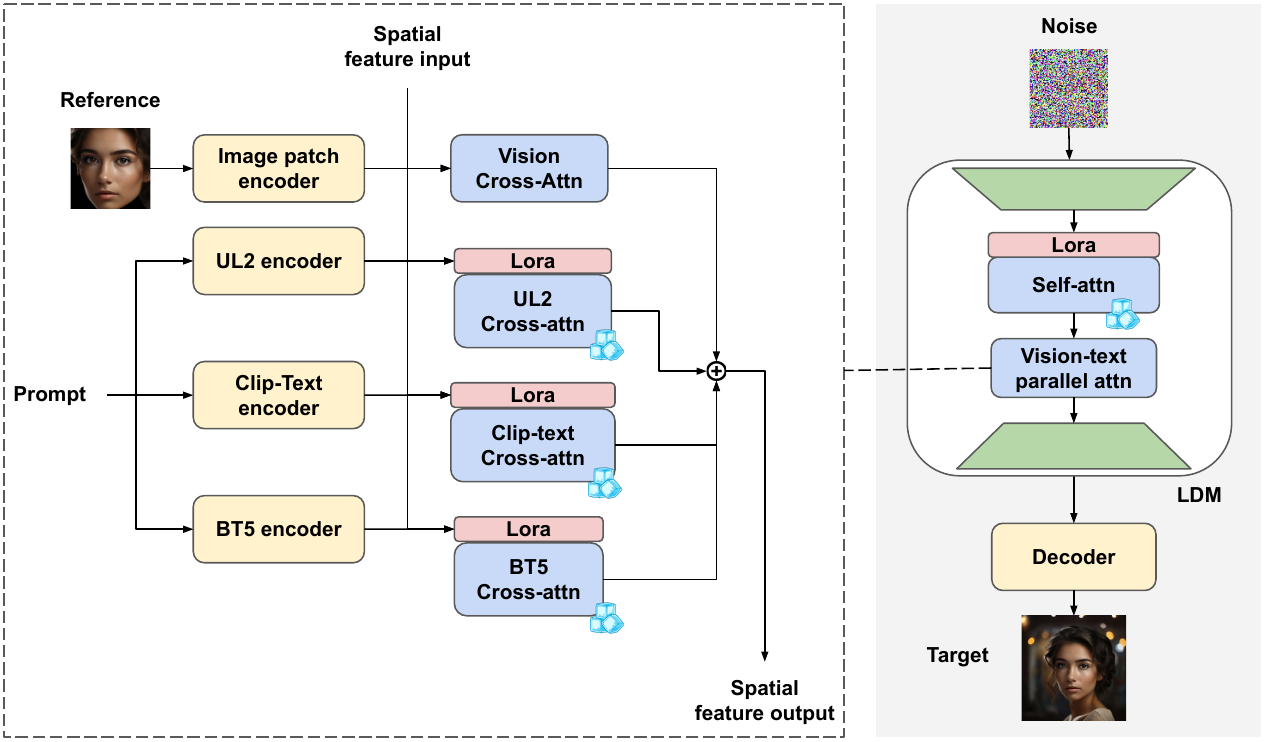}
\caption{Overview of \memu model architecture. We introduced a fully parallel architecture
that incorporates three text encoders and a trainable vision encoder for optimizing identity preservation and text-alignment. We adopted LoRA on top of the self-attention layers and the text cross-attention layers to best preserve the foundation model's image generation quality.} \label{img:method:overview}
\end{figure}
% ------------------------------------------

\subsection{Synthetic Paired Data (\texttt{SynPairs})} \label{sec:method:synpair}

%(\animesh{It feels sudden to introduce SynPairs here, I think we should talk about limitations of unpaired data before showing how we fixed it? As a reader, I'd think why use large scale SynPairs}) 

We observed that one critical issue during training is the use of unpaired data, \ie, the cropped image as input and the original image as target. It can introduce a severe copy-paste effect, making the model hard to learn the true identity relationship between input and output more than duplicating the reference image. Thus, the model is not able to generate images that follow hard prompts, \eg, change expression or head orientation.

To this end, we proposed a new synthetic data generation recipe to create high-quality paired data (same identity with varying expression, pose, and lighting conditions, \etc.) for training. Compared to directly sourcing real paired data, which is not readily available, our study shows that curating the paired data synthetically allows us to retain higher quality data to further enhancing several aspects of the \memu model.

% ------------------------------------------
\begin{figure}[h]
\centering
\includegraphics[width=\linewidth]{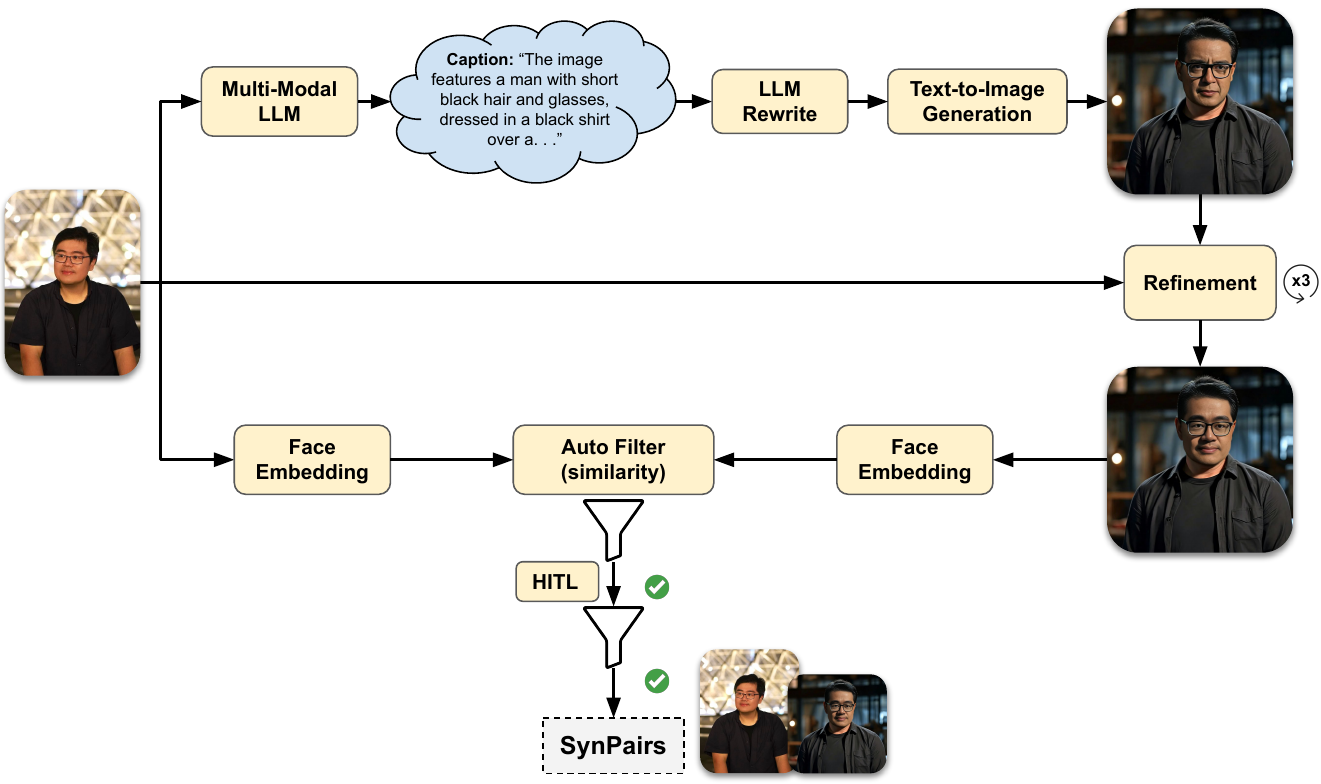}
\caption{Generation pipeline for \texttt{SynPairs} data. We first caption real images using multi-modal LLM and rewrite through a LLM rewriter. The prompt is fed into a text-to-image generation model to obtain high-quality synthetic images, and then refined with the reference image to better preserve identity. This results in high-quality paired data, \ie, same identity with varying expression, pose, and lighting conditions, \etc.}
\label{img:method:synpair}
\end{figure}
% ------------------------------------------

To generate \texttt{SynPairs} data, we first obtain a dense image caption of the real reference image via a multi-modal LLM. The caption then flows through a caption rewrite stage based on Llama3 \citep{meta_llama3_blogpost} to inject more gaze and pose diversity in the caption. The rewritten caption is then fed to a text-to-image generation tool such as Emu \citep{dai2023emu} as the prompt to produce a high quality synthetic images. Next, we refine the generated image identity based on the reference image identity. After a large number of curated synthetic pairs is generated they go through an automatic filter based on similarity.

\subsection{Model Architecture} \label{sec:method:arch}

\subsubsection{Vision Encoder} \label{sec:method:arch:vision_encoder}

We propose to use a trainable CLIP ViT-H patch vision encoder to extract the identity control signal from the reference image.
%(\animesh{Why CLIP, why not DinoV2? Maybe we need an ablation to show the comparison} \zechengh{low priority to ablation this given the timeline}). 
Unlike previous work that heavily relied on face embedding, we observed that a general trainable vision encoder can provide adequate information to preserve the identity.

To further improve the identity preservation capability, we crop the face area and mask the corresponding background of the reference image to avoid the model attending to the non-critical areas, \eg, image background and non-face area in the cropped image. Figure 2 illustrates the vision embedding workflow. We also proposed to use \texttt{zero\_conv} as initialization to avoid adding noisy control signals at the beginning of training.

\subsubsection{Text Encoders} \label{sec:method:arch:text_encoder}

We employ three distinct text-encoders: CLIP ViT-L \citep{radford2021learning} text encoder, UL2 \citep{tay2022ul2}, and ByT5 \citep{xue2022byt5}, as the text conditioning mechanisms. The selection of these encoders is driven by their respective strengths and suitability for specific tasks. The CLIP text encoder, for instance, shares a common space with the CLIP vision encoder, facilitating enhanced identity preservation. To capitalize on this alignment, we initialize the cross-attention module of the vision encoder with the pre-trained CLIP text encoder. %(\animesh{In the above section, we said randomly initialized vision cross attention module?} \zechengh{Due to vision and text embedding are not well-aligned at the beginning}). 
Meanwhile, UL2 is specifically chosen for its proficiency in comprehending long and intricate text prompts, making it instrumental in handling complex input data. Furthermore, the ByT5 model is integrated for its supreme capability in encoding characters. We leverage ByT5 to improve visual text generating in the image, \eg, text on a signage.

\subsubsection{Fully Parallel Image-Text Fusion} \label{sec:method:arch:parallel_attn}

We investigated a parallel attention architecture to incorporate the vision and text conditions. Specifically, the newly added vision condition from the reference image and the spatial features fuse through a new vision cross-attention module. The output of the new vision cross-attention module is then added to the text cross-attention output. In our experiments, this design better balances the vision and text control than concatenating the text and vision controls. %(\animesh{Do we have numbers to show this ablation? I think we might need a detailed ablation table on text-image fusion methods. Like concat, parallel, etc}).
%-----------------------------------
\begin{figure}[t]
    \centering
    \includegraphics[width=\linewidth]{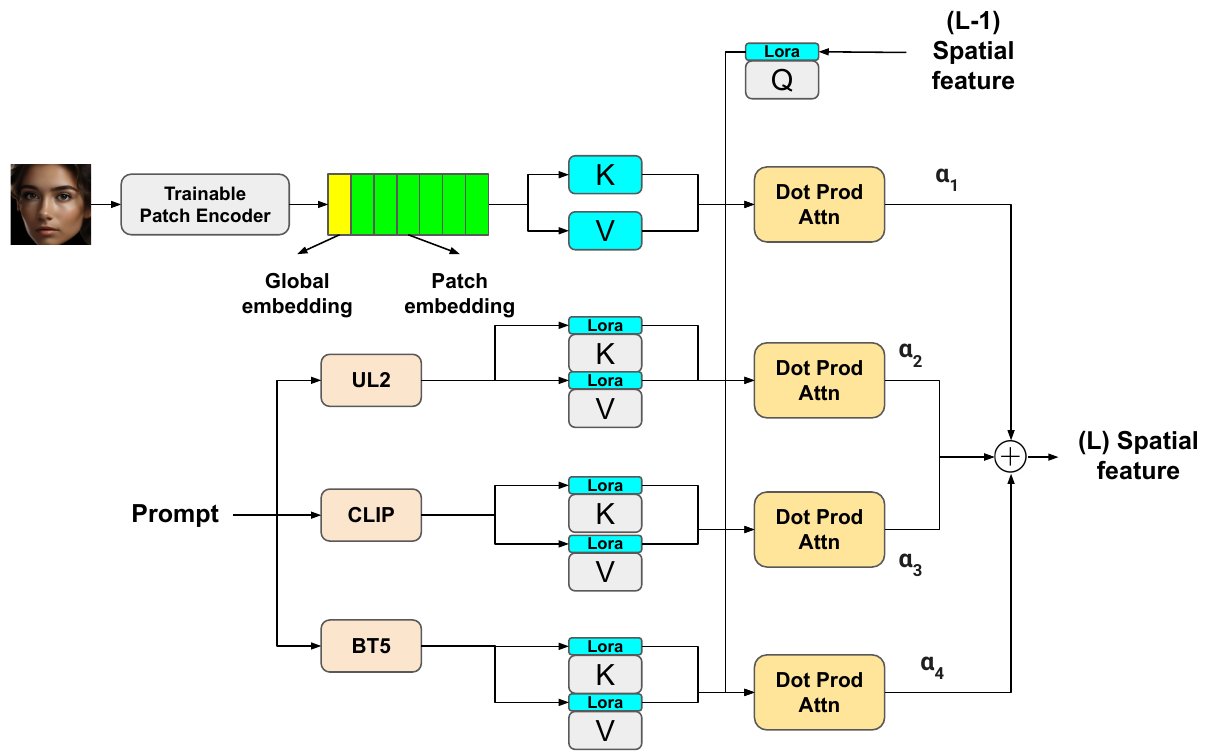}
    \caption{Fully parallel image-text fusion architecture. We employ three distinct text-encoders: CLIP ViT-L \citep{radford2021learning} text encoder, UL2 \citep{raffel2020exploring}, and ByT5 \citep{xue2022byt5}, as the text conditioning. They interact with a trainable CLIP vision encoder through fully parallel attention fusion.} \label{img:method:parallel_attn}
\end{figure}
%-----------------------------------

\subsubsection{LoRA} \label{sec:method:arch:lora}

To preserve the visual quality from the foundation model, we leveraged low-rank adapters (LoRA) on top of the cross-attention module. The self-attention and text cross-attention modules in the foundation Unet are frozen. We observed that this design not only better preserves the foundation model’s image generation capability, but also accelerates the convergence speed by up to 5x. % (\animesh{A question to ask here is why don't we apply Lora on vision cross attention layers too after 1st stage pretraining?}.})

\subsection{Multi-Stage Finetune} \label{sec:method:finetune}

\begin{figure}[h]
    \centering
    \includegraphics[width=1.0\linewidth]{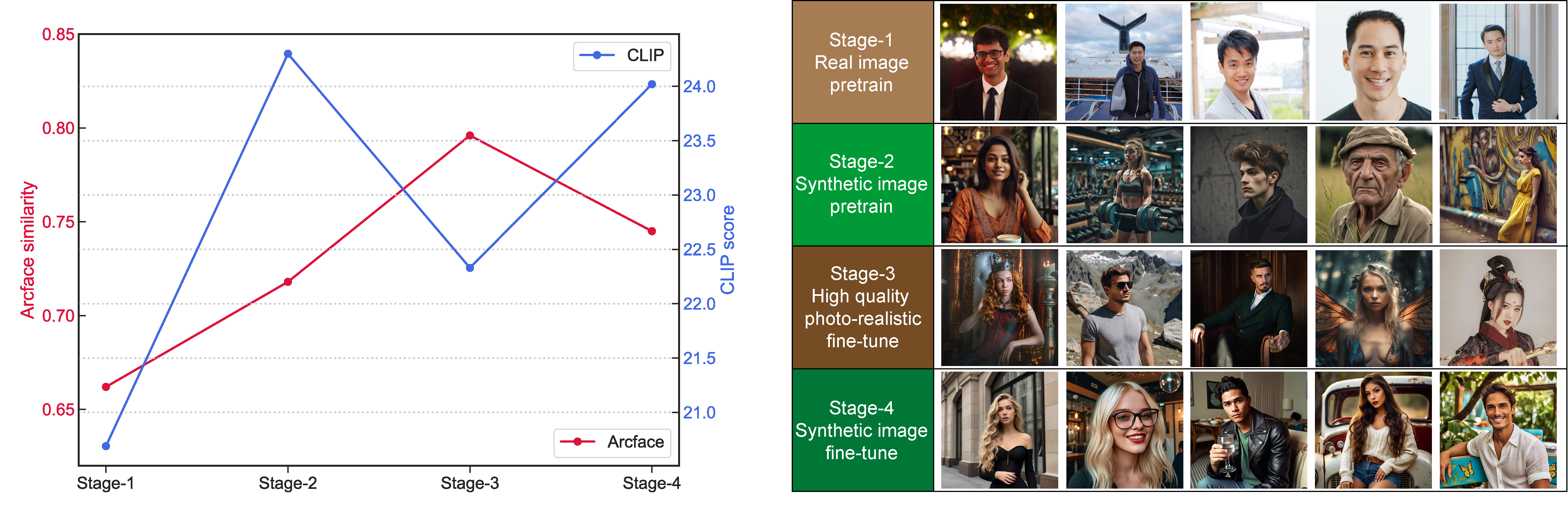}
    \caption{Training with real images has higher identity, training with synthetic images has higher prompt alignment. After an interleaved multi-staged training, identity and prompt alignment achieves best trade-off.}
    \label{fig:multi_stage}
\end{figure}

We propose a multi-stage finetuning with interleaved real and synthetic data that help us achieve the best trade-off between editability and identity preservation. In the first two stages, we leverage large-scale data (nine millions) to pretrain the model to be able to condition on a reference identity. For the later stages, we finetune our pretrained checkpoint with high-quality, aesthetic images collected through Human-In-The-Loop (HITL). Empirically, we found training with real images gives the best identity preservation, while training with synthetic images gives better prompt alignment (editability). Synthetic images are generated from its respective prompt, therefore the image-text alignment is high and there is less noisy information during training, but the identity information is not as rich as in real data. This is why we adopt the interleaved training recipe, as shown in Figure \ref{fig:multi_stage}. After the 1st real data pretraining, the model is able to condition on image, after 2nd synthetic data pretraining, the prompt alignment is high but identity is not perfect, after the 3rd high-quality real-data finetune, the identity is good but prompt alignment drops, the 4th high-quality synthetic data finetune achieves the best trade-off between identity and editability. % (\animesh{A question that comes up here is why not do 1st stage real pretraining, then directly do 4th stage synthetic finetuning rather than improving prompt alignment, then dropping it, then improving it again.}) (\animesh{So here we don't talk about SynPairs, because that wasn't need to get to sota right?}) (\animesh{We haven't mentioned in the whole section that the input and target images come from the same image, because that might make extension to SynPairs more clearer later})

% During model training, we proposed a multi-stage fine-tune methodology where the data quality is gradually improved. In the earlier two stages, we leveraged auto-filtered data to train the model being able to follow the identity condition signals, while we used HITL in the later two stages to further improve the visual quality of the generated images.

% To further improve the visual appeal of the generated images, we proposed multi-stage finetuning. We use higher-quality and fewer data in later stages than the earlier stages. Specifically, for training the Emu-Personalization model, we leveraged four stages. The first two stages leveraged a large amount of auto-filtered real and synthetic images, respectively, to pretrain the model being able to follow the identity control signals. The later two stages are tuned with high-quality HITL filtered data. We observed that the multistage finetuning improves both visual appealing and text-alignment of the model. Figure 4 shows the four training stages and exemplary training data from each stage.

\subsection{Extension to Multi-Subject Personalization} \label{sec:method:multi_people}

The previously introduced fully parallel image-text fusion pipeline (Section~\ref{sec:method:arch:parallel_attn}) can be flexibly extended to accommodate multi-subject personalization. In the two-person scenario for example, instead of passing the global embedding and patch embedding of the single reference image into the $\mathbf{K}$ and $\mathbf{V}$ components as shown in the top left branch of Figure~\ref{img:method:parallel_attn}, we can concatenate the vision embedding from both reference images and passing it into the $\mathbf{K}$ and $\mathbf{V}$ components. Given this setup, through training, the network learns how to map from reference$_i$ to subject$_j$ in the group photo while generating prompt-induced image context accordingly. Some examples of the two-person personalization results are shown in Figure~\ref{img:exp:two_person}.

\section{Experiments} \label{sec:exp}

In this section, we perform both qualitative and quantitative evaluations of our model. We also compare our model to the SOTA personalization models. Results show that our model outperforms the existing models on all axes setting the new state-of-the-art.

% \subsection{Experiment Settings} \label{sec:exp:setting}

\subsection{Qualitative Evaluation} \label{sec:exp:qualitative}

We show examples of our model generated image in Figures \ref{img:exp:shine_1}-\ref{img:exp:shine_5}. Our model generates visually appealing images that both preserve the identity and follow the prompt faithfully.

%-----------------------------------
\begin{figure}[t]
    \centering
    \includegraphics[width=\linewidth]{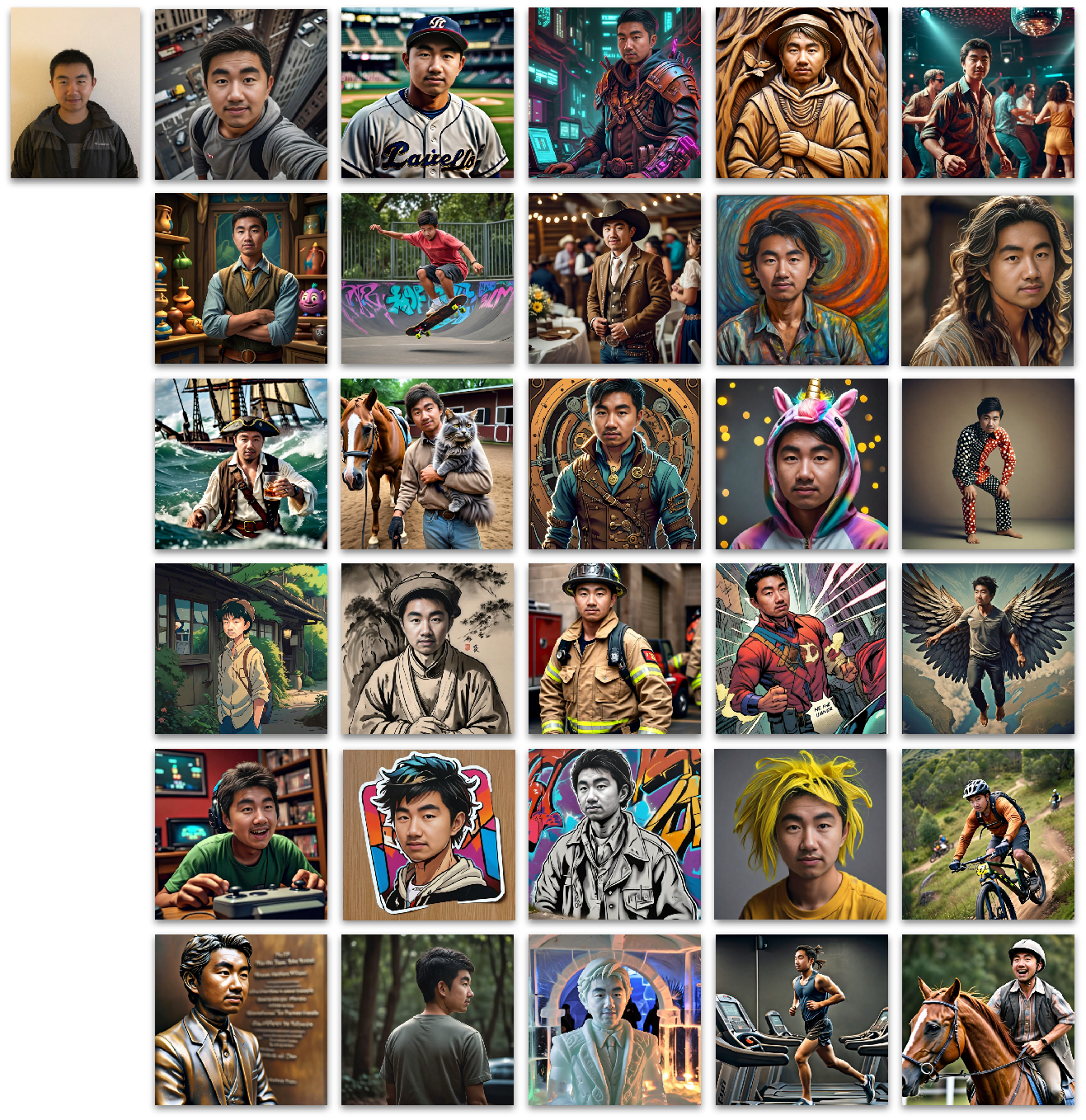}
    \caption{More visualizations (1/5) of generated personalized images using \memu.} \label{img:exp:shine_1}
\end{figure}
%-----------------------------------
%-----------------------------------
\begin{figure}[t]
    \centering
    \includegraphics[width=\linewidth]{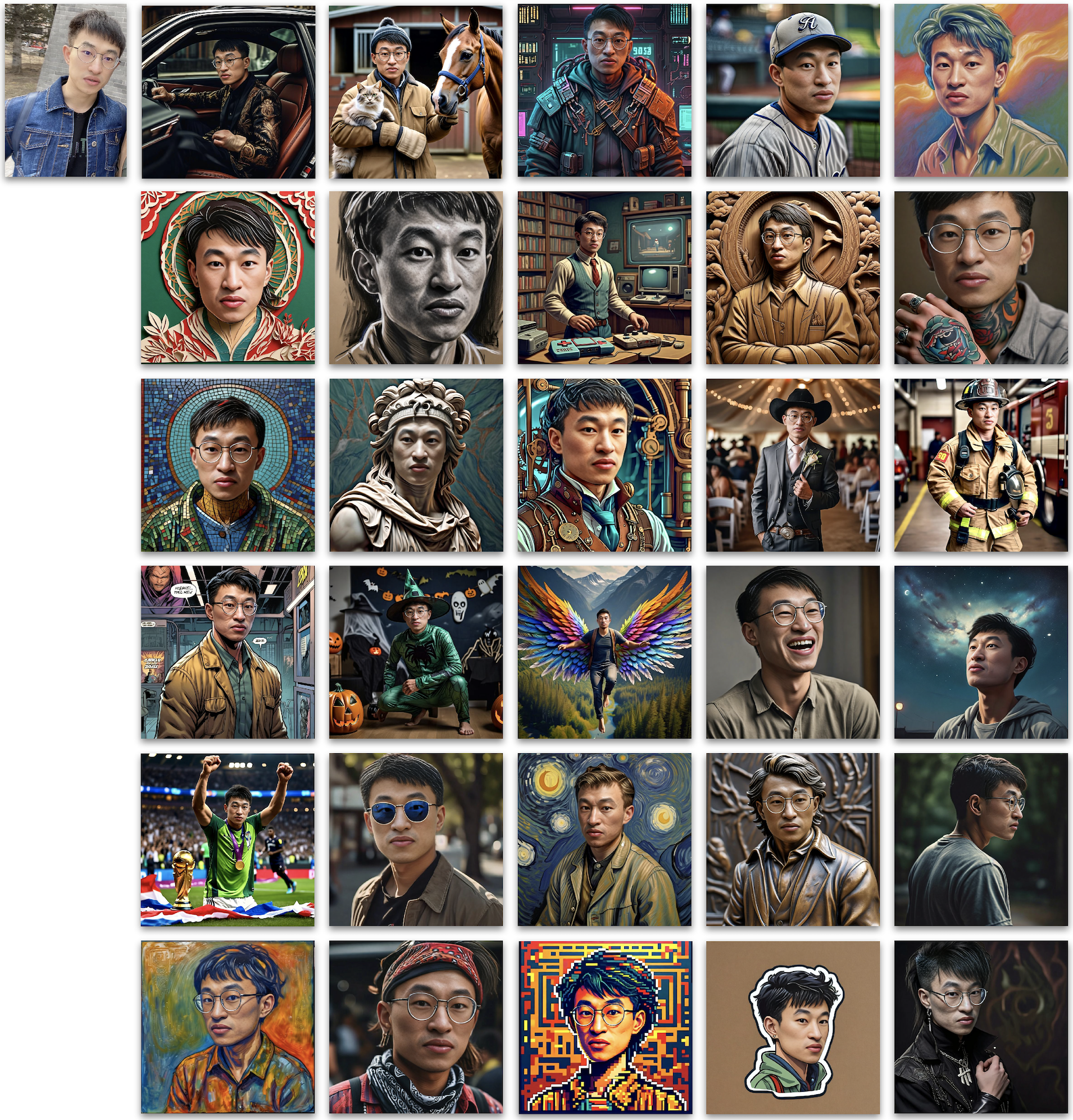}
    \caption{More visualizations (2/5) of generated personalized images using \memu.} \label{img:exp:shine_2}
\end{figure}
%-----------------------------------
%-----------------------------------
\begin{figure}[t]
    \centering
    \includegraphics[width=\linewidth]{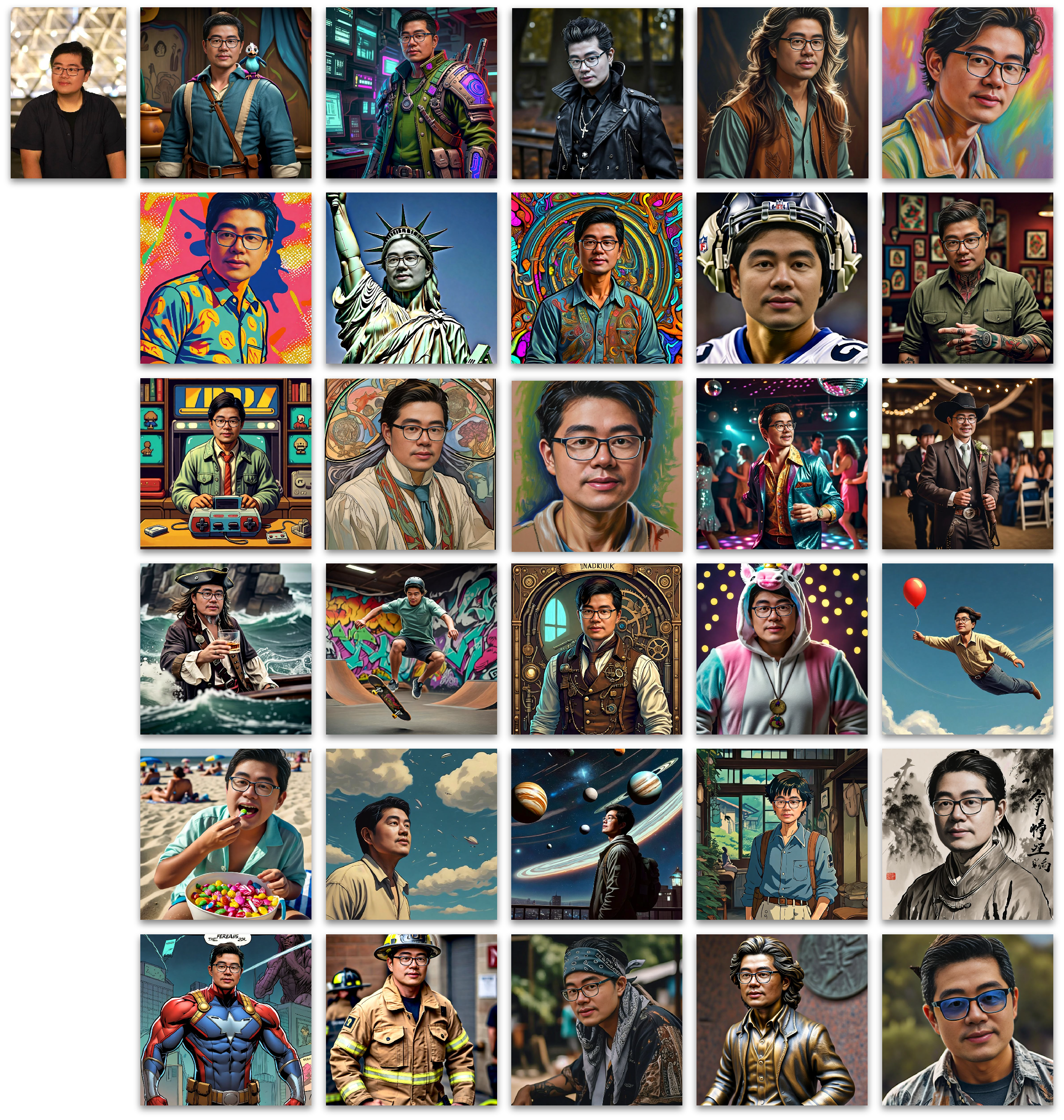}
    \caption{More visualizations (3/5) of generated personalized images using \memu.} \label{img:exp:shine_3}
\end{figure}
%-----------------------------------
%-----------------------------------
\begin{figure}[t]
    \centering
    \includegraphics[width=\linewidth]{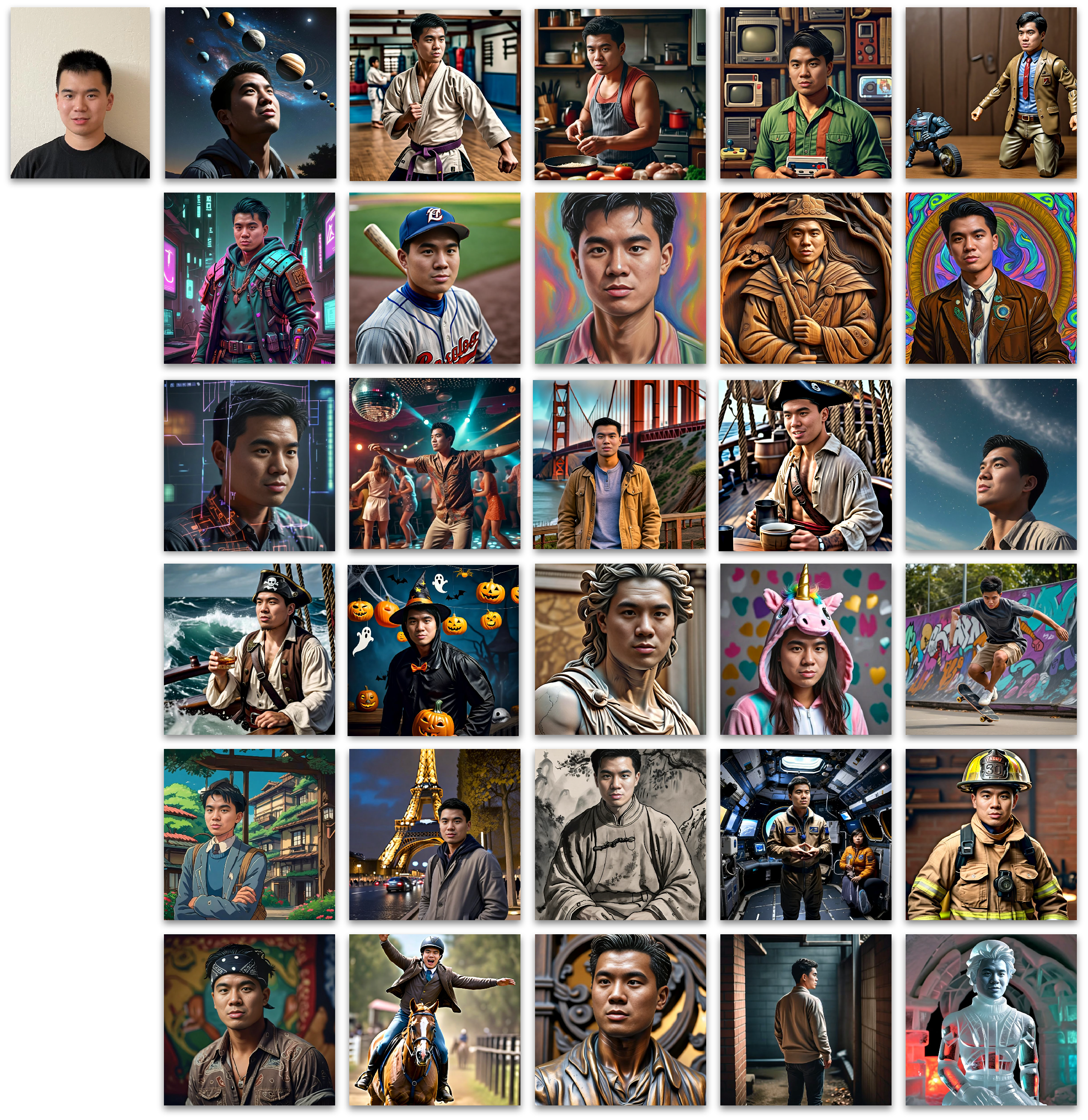}
    \caption{More visualizations (4/5) of generated personalized images using \memu.} \label{img:exp:shine_4}
\end{figure}
%-----------------------------------
%-----------------------------------
\begin{figure}[t]
    \centering
    \includegraphics[width=\linewidth]{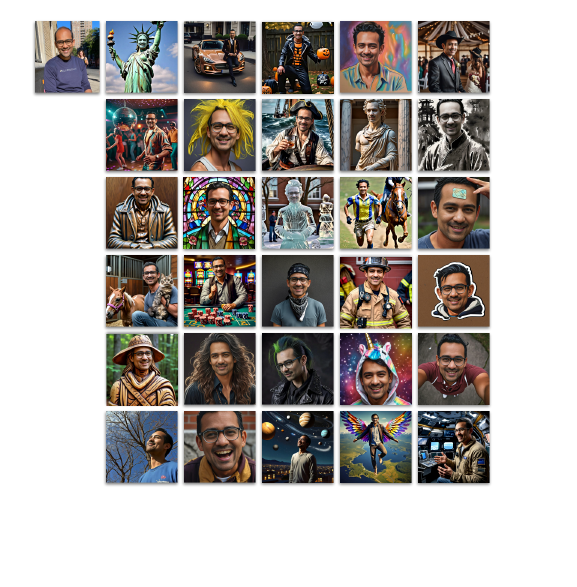}
    \caption{More visualizations (5/5) of generated personalized images using \memu.} \label{img:exp:shine_5}
\end{figure}
%-----------------------------------

\begin{figure}[t]
    \centering
    \includegraphics[width=\linewidth]{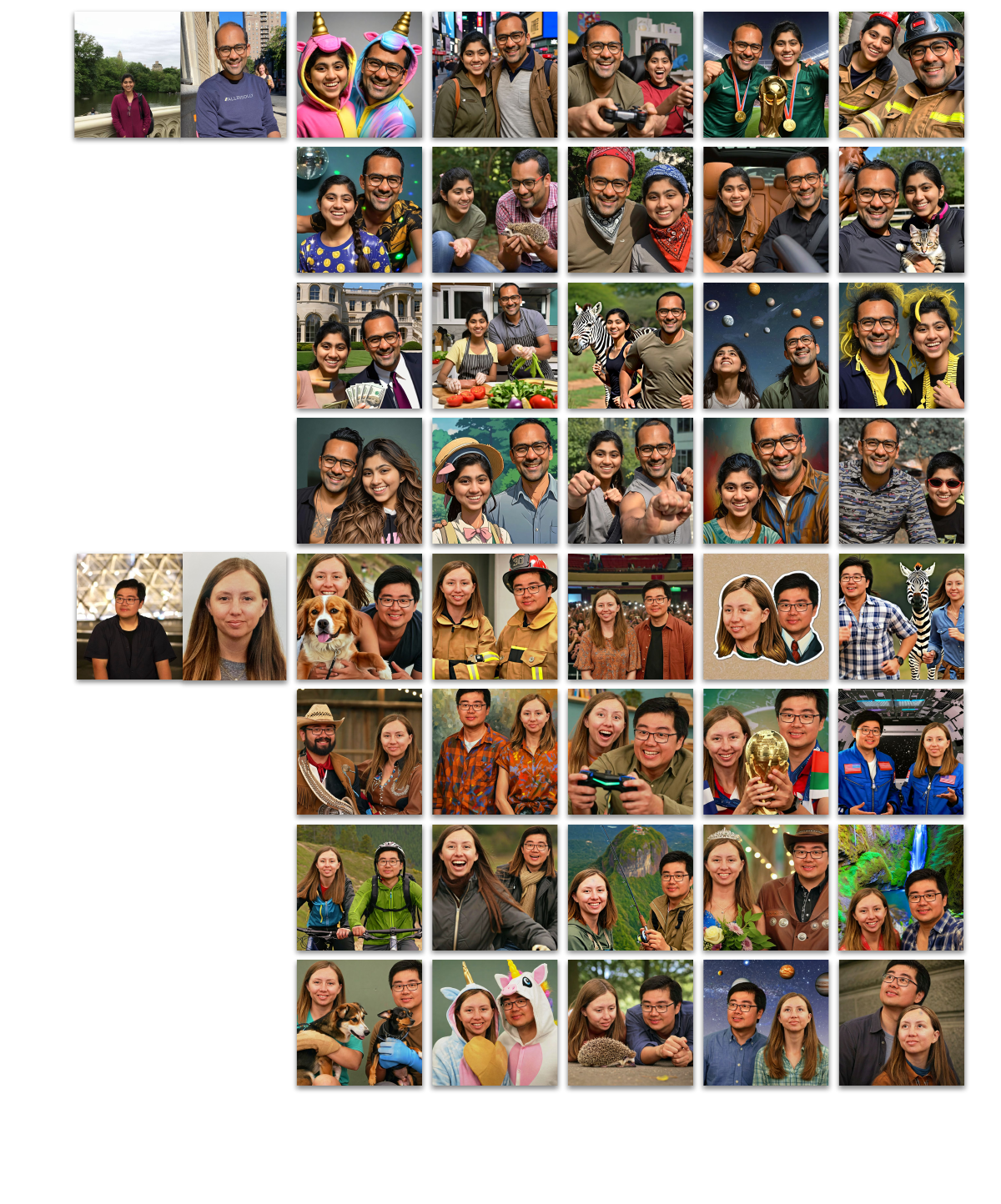}
    \caption{Multi-player personalization using \memu.} \label{img:exp:two_person}
\end{figure}

\subsection{Quantitative Evaluation} \label{sec:exp:quantitative}

\subsubsection{Evaluation Dataset}

% To quantitatively evaluate \memu, we created an evaluation set consisting of two parts: (i) reference images, and (ii) eval prompts. To have a comprehensive comparison in all representative cases, we collected a total of 51 reference identities covering different gender, race, and skin tone: 1) real user images, 2) celebrity images, 3) external benchmarks, 4) synthetic identities. We created a list of 65 prompts to evaluate the model, including hard prompts that require face expression or pose changes. Distributions of the reference images and the prompts are shown in Figure~\ref{img:exp:quantitative:distribution}.

To quantitatively evaluate \memu, we created an evaluation set consisting of two parts: (i) reference images, and (ii) eval prompts. To have a comprehensive comparison in all representative cases, we collected a total of 51 reference identities covering different gender, race, and skin tone. We created a list of 65 prompts to evaluate the model. It widely covers a wide range of usage scenarios, and also including hard prompts that require face expression or pose changes, camera motions, and stylization. These prompts help to assess the model's ability to engage in more complex and nuanced interactions, diverse pose generation, and harmonization. Each identity is paired with all 65 prompts, so a total of 51x65=3315 generations for one round of human evaluation. The distributions of the prompts is shown in Figure~\ref{fig:heval_dist}.

\begin{figure}[H]
    \centering
    \includegraphics[width=0.85\linewidth]{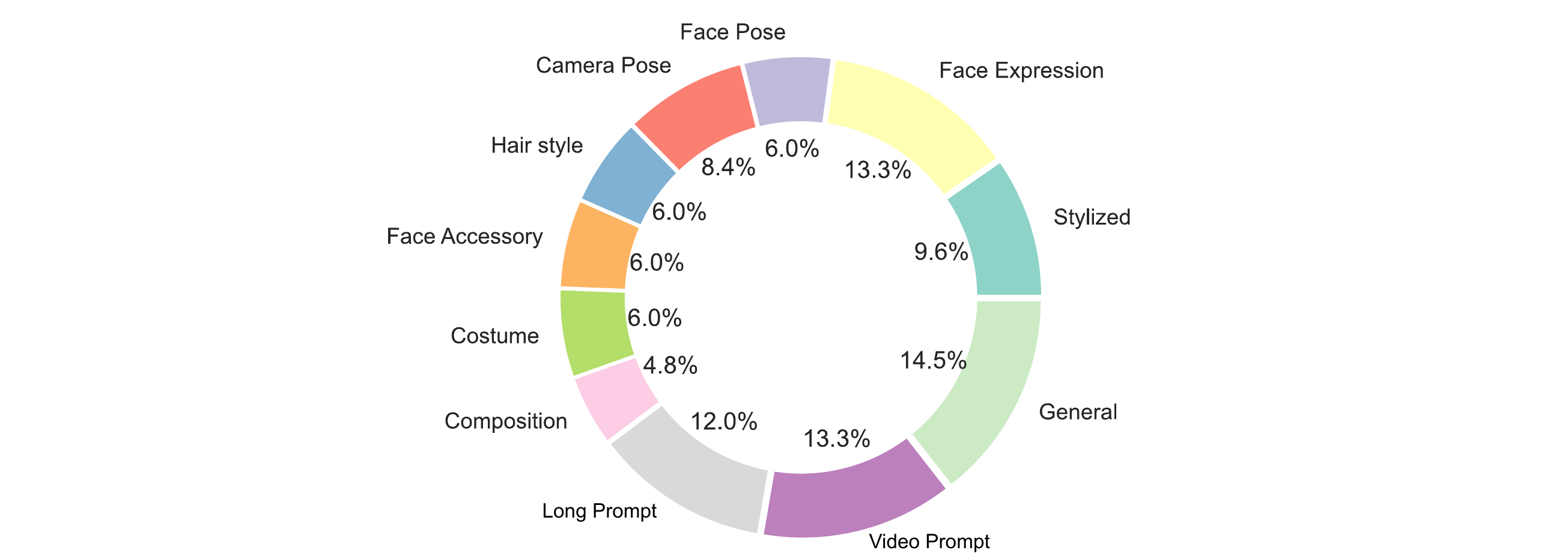}
    \caption{Distribution of the evaluated prompts. It widely covers a wide range of usage scenarios, and also including hard prompts that require face expression or pose changes, camera motions, etc.}
    \label{fig:heval_dist}
\end{figure}

% DreamBooth \citep{ruiz2023dreambooth} is the only tuning-based method shown in the figure. For fair comparison, we finetune the same foundation model with the single reference image of each subject and $100$ prior-preservation images. More specifically, we use LoRA \citep{hu2021lora} with a rank of $4$ to finetune the model for $1000$ iterations and with a learning rate of $10^{-4}$.

\subsubsection{Benchmarked Methods}

We benchmarked the SOTA adapter-based personalization model and the SOTA control-based model. For the adapter-based method, we select the best one that strikes the best balance among visual appeal, identity preservation, and prompt alignment, the three axes that we evaluate our models on. For control-based method, we noticed that the choice of the pose image plays an important role in how the final generated image is composed, \ie, for some prompts, a carefully chosen pose image can make the generated images look better or worse. For a fair comparison, we use the reference image itself as the pose condition.

\subsubsection{Human Evaluation}

% To quantitatively evaluate the generated images, we established a large-scale annotation process that assesses various quality axes of the generated images. We use human annotation as the golden standard to assess the model’s quality (standalone evaluation) and compare between models (head-to-head evaluation).

% In the standalone evaluation, we show annotators the input image, prompt and a generated image and ask them three questions to rate on a scale of Strong Pass / Weak Pass / Fail. (1) \textbf{Identity Similarity}: Given the subject in the original image, does the subject in the output image appear to have the same identity? (2) \textbf{Prompt Alignment}: Given the original image, does the output image follow the personalization prompt faithfully? (3) \textbf{Visual Appeal}: Is the output image visually appealing? In the head-to-head model evaluation, we compare one model against another on the same three axes discussed above.

% As shown in Table~\ref{tab:hev_h2h_x2}, our proposed \memu outperforms the two SOTA methods IP-Adapter-Face-Plus (IPA) and Instant-ID by a margin in most axis. Specifically, \memu is significantly better in prompt alignment, +45.1\% and +30.8\% against InstantID and IP-Adapter-Face-Plus, respectively. One exception is Instant-ID is better in identity preservation than \memu. We observed that it is because InstantID hard copy-pastes the reference image at the center of the image, resulting in unnatural images though the identity metric is high. 

To evaluate the quality of the generated images, we conducted a large-scale annotation process that assessed various aspects of the images. We used human annotation as the gold standard to assess the model's performance (standalone evaluation) and compare it with other models (head-to-head evaluation).

In the standalone evaluation, we presented annotators with the input image, prompt, and a generated image and asked them three questions to rate on a scale of Strong Pass / Weak Pass / Fail. (1) \textbf{Identity Similarity}: Does the subject in the output image appear to have the same identity as the subject in the original image? (2) \textbf{Prompt Alignment}: Does the output image follow the personalization prompt faithfully? (3) \textbf{Visual Appeal}: Is the output image visually appealing?
In the head-to-head model evaluation, we compared one model against another on the same three axes.

As shown in Table~\ref{tab:hev_h2h_x2}, \memu outperforms the two state-of-the-art methods adapter-based model and control-based model by a significant margin in most axes. Specifically, \memu is significantly better in prompt alignment, with a +45.1\% and +30.8\% improvement over the SOTA adapter-based model and the SOTA control-based model, respectively. However, we observed that the control-based model is better in identity preservation than \memu, due to its hard copy-pasting of the reference image at the center of the image, resulting in unnatural images despite the high identity metric.

% ------------------------------------------
% \begin{table}[htbp]
%   \centering
%   \scriptsize
%   \caption{Quantitative evaluation results of \memu vs. IPA-FaceID-Plus under both standalone and head-to-head human evaluation settings.}
%   \begin{tabular}{c|cc|ccc}
%     \toprule
%     ~ & \multicolumn{2}{c}{\textbf{Standalone} (strong pass + weak pass rate)} & \multicolumn{3}{c}{\textbf{Head-to-head} (win rate)} \\ \midrule
%     \textbf{Metrics}        & \textbf{IPA-FaceID-Plus} & \textbf{\memu} & \textbf{IPA-FaceID-Plus} & \textbf{\memu} & \textbf{Tie} \\ \midrule
%     Prompt Alignment        & 61.2\% & \textbf{86.5\%} & 1.6\% & \textbf{32.4\%} & 66.0\% \\
%     Identity Preservation   & 94.1\% & \textbf{97.9\%} & 3.8\% & \textbf{5.5}\%  & 90.7\% \\
%     Visual Appeal           & 96.7\% & 96.5\%          & 3.3\% & \textbf{4.2}\%  & 92.5\% \\
%     \bottomrule
%   \end{tabular} \label{tab:hev}
% \end{table}
% ------------------------------------------

\subsection{Ablation Study} \label{sec:exp:ablation}

In our ablation study, we examined the effectiveness of various components within our proposed \memu. Main ablation results are shown in Table \ref{tab:exp:ablation}.

\begin{table}[htbp]
  \centering
  \scriptsize
  \caption{Ablation study of different component in \memu.}
  \begin{tabular}{c|c|c|c}
    \toprule
    ~ & \multicolumn{3}{c}{\textbf{Standalone Pass Rate ($\uparrow$)}} \\ \midrule
    \textbf{Model}                     & Prompt Alignment    & Identity Preservation & Visual Appeal  \\ \midrule
    \memu w/o multi-stage finetune     &  55.3\%  & 90.7\%   & 45.7\% \\ \midrule
    \memu w/o fully parallel attention & 75.6\%  & 81.9\%   & 65.7\%  \\ \midrule
    \memu w/o \texttt{SynPairs}                 &  57.0\% & \textbf{95.5}\%  & 83.4\% \\ \midrule
    \memu  & \textbf{80.8} \%  & 83.3\%  &  \textbf{87.7}\% \\
    \bottomrule
  \end{tabular} \label{tab:exp:ablation}
\end{table}

\subsubsection{Impact of Multi-stage Finetune}

The ablation results highlight the impact of multi-stage fine-tuning. Reducing the multi-stage fine-tuning to a single stage significantly degrade all metrics, especially 25.5\% in prompt alignment and 42.0\% in visual appeal. Moreover, we observe that, the synthetic fine-tune stages provides better prompt alignment and the real data fine-tune stage improves the identity preservation capability.

\subsubsection{Impact of Fully Parallel Attention}

We ablate removing the full parallel attention to a standard token concatenate design to show the impact of the fully parallel attention architecture. We observed that all metrics, specifically 5.2\% in prompt alignment, 1.4\% in identity preservation, and 22.0\% in visual appeal, respectively. This shows the importance incorporating all three text encoders and the vision encoder through fully parallel attention.

\subsubsection{Impact of Synthetic Pairs}

\texttt{SynPairs} increase the diversity of the generated images by eliminating the copy-paste effect. Our ablation verifies this assumption and demonstrate better prompt-alignment compared to the model without synthetic paired training. We observed that it is especially effective for the complex prompts that require strong changes to the original images, \eg, expression change, covering the face, or turning head, \etc. However, we observed a regression in identity preservation with \texttt{SynPair} training because the faces in the corresponding reference and target pair are not exactly the same. Future work will focus on improving the face similarity of \texttt{SynPair} training data.

\section{Future Work}

We would like to continue research and explore the following directions: 1) extend personalized image to video generation. The key is to consistently preserve the identity and scene in video generation. 2) While \memu has improved the prompt-alignment against existing models, we observed that it still has limitation in following prompts describing very complex poses, \eg, \textit{jumping from a mountain}. Future work will focus on improving the generated images' quality on these prompts.

\section{Conclusion} \label{sec:conclusion}

In this study, we introduce \memu, a pioneering model tailored for personalized image generation. Unlike traditional tuning-based approaches, \memu operates as a tuning-free solution, offering a shared framework accessible to all users without the need for individual adjustments. \memu overcomes the prior research limit in handling the intricate balance between preserving identity, following complex prompts, and maintaining visual quality by introducing 1) a novel synthetic paired data generation mechanism to foster image diversity, 2) a fully parallel attention architecture featuring three text encoders and a fully trainable vision encoder to enhance text faithfulness, and 3) a novel coarse-to-fine multi-stage fine-tuning methodology to progressively enhance visual quality. We perform large-scale human evaluation on thousands of examples and showcase that \memu outperforms state-of-the-art personalization models, demonstrating superior capabilities in identity preservation, visual quality, and text alignment.

\section*{Acknowledgment}

We extend our gratitude to the following people for their contributions: Xiaoliang Dai, Ji Hou, Kevin Chih-Yao Ma, Kunpeng Li, Sam Tsai, Jialiang Wang, Matthew Yu, Simran Motwani, Zijian He for text-to-image foundation model; Eric Alamillo, Xiao Chu, Yangwen Liu, Yan Yan, Jonas Kohler, Artsiom Sanakoyeu, Ali Thabet, Arantxa Casanova Paga, Zhipeng Fan for model evaluation and discussion; Aaron Nissenbaum, Becky McMahon, Mo Metanat, Danny Trinh, Jack Hanlon, Tali Zvi, Manohar Paluri, Prashant Ratanchandani, and Ahmad Al-Dahle for support and leadership.

\bibliographystyle{assets/plainnat}
\bibliography{ref}

\end{document}